# Design, Development, and Use of Maya Robot as an Assistant for the Therapy/Education of Children with Cancer: a Pilot Study

**Alireza Taheri[1*], Minoo Alemi[1,2], Elham Ranjkar[2], Raman Rafatnejad[1], Ali F. Meghdari[1,3]**

[1] Social & Cognitive Robotics Laboratory, Center of Excellence in Design, Robotics, and Automation (CEDRA), Sharif University of Technology, Tehran, Iran

[2] West Tehran Branch, Islamic Azad University, Tehran, Iran

[3] Chancellor, Fereshtegaan International Branch, Islamic Azad University, Tehran, Iran

* Corresponding Author: artaheri@sharif.edu ; +982166165531

**ABSTRACT** This study centers around the design and implementation of the Maya Robot, a portable elephant-shaped social robot, intended to engage with children undergoing cancer treatment. Initial efforts were devoted to enhancing the robot's facial expression recognition accuracy, achieving a 98% accuracy through deep neural networks. Two subsequent preliminary exploratory experiments were designed to advance the study's objectives. The first experiment aimed to compare pain levels experienced by children during the injection process, with and without the presence of the Maya robot. Twenty-five children, aged 4 to 9, undergoing cancer treatment participated in this counterbalanced study. The paired T-test results revealed a significant reduction in perceived pain when the robot was actively present in the injection room. The second experiment sought to assess perspectives of hospitalized children and their mothers during engagement with Maya through a game. Forty participants, including 20 children aged 4 to 9 and their mothers, were involved. Post Human-Maya Interactions, UTAUT questionnaire results indicated that children experienced significantly less anxiety than their parents during the interaction and game play. Notably, children exhibited higher trust levels in both the robot and the games, presenting a statistically significant difference in trust levels compared to their parents (P-value < 0.05). This preliminary exploratory study highlights the positive impact of utilizing Maya as an assistant for therapy/education in a clinical setting, particularly benefiting children undergoing cancer treatment. The findings underscore the potential of social robots in pediatric healthcare contexts, emphasizing improved pain management and emotional well-being among young patients.

## I. INTRODUCTION

*A. Motivation of the study*

Social robots serve as intelligent companions, engaging with humans in various aspects of their daily lives, enhancing performance, and contributing to overall well-being [1-3]. Specifically designed to assist in education, healthcare, and domestic environments, these robots play a crucial role in augmenting human experiences and providing social services [1-3]. Their interactive capabilities enable expanded social engagement, particularly with children, fostering advanced medical treatments, enriching educational experiences, enhancing social skills, and supporting the mental health of children, especially those undergoing cancer treatment [4, 5]. Children, naturally drawn to moving and smart toys, find particular interest in robots, offering a valuable avenue to alleviate anxiety, improve educational outcomes, and provide entertainment, especially in healthcare settings [6-8]. Recognizing the importance of addressing both the emotional and physical well-being of children in hospital settings, the design of the Maya Robot is purposefully tailored to function as a companion and toy friend. Maya's role is not merely utilitarian; it extends to alleviating pain and anxiety, serving as a source of comfort for children undergoing medical treatments. Through this approach, Maya seeks to contribute significantly to the emotional support and overall well-being of children, recognizing the unique impact that interactive technologies, particularly social robots, can have on pediatric healthcare experiences.

Given the unique circumstances surrounding children with cancer and the constraints on accessing them within specific time frames, obtaining additional volunteers and investing extended periods to enhance examination accuracy proves impractical. In light of these challenges, the robot serves a dual purpose by not only aiding in examinations but also functioning as a non-human source of entertainment and education. This multifaceted role becomes particularly crucial in mitigating the risk of infectious disease transmission from treatment staff to the child, given the diminished immunity of children undergoing cancer treatment. Recognizing the limited access to volunteers and the time constraints in the context of pediatric cancer care, the robot's capacity to serve as an educational and entertainment tool becomes paramount. This not only addresses the challenges posed by restricted access but also contributes significantly to infection control measures. By minimizing direct human interaction, especially from treatment staff, the robot acts as a protective barrier, reducing the potential transmission of infections to children who are already immunocompromised due to cancer treatments. In essence, the robot's role extends beyond examinations, playing a vital part in safeguarding the health of these vulnerable children and enhancing the overall quality of care in pediatric oncology settings. In addition, it should be considered that children with cancer experience a (physically and mentally) painful process during their treatment in the pediatric hospitals. Contrary to their healthy peers, they also might face the lack of appropriate educations because of being away from the regular educational environments. The mentioned issues show the serious needs to think about providing an appropriate game-based environment (especially based on modern technologies such as social robotics) in which children with cancer (and their families) can experience a calmer place with potential of learning required materials/skills.

In this paper, we present our developed robot-assisted game-based package (equipped with AI) to be used in pediatric hospitals as well as the results of conducting this package for some children with cancer during a preliminary exploratory study.

*B. Related Works and Literature Review*

Numerous studies have been conducted to design, analyze, and enhance different aspects of social robots deployed in medical settings. In a comprehensive review by Dawe et al. [9], it was noted that robots have the potential to enhance children's medical knowledge by providing relevant information in healthcare settings. Van Bindsbergen et al. [10] investigated the application of a NAO robot in sleep hygiene education for 28 children and their parents at a pediatric oncology outpatient clinic. The study aimed to assess the feasibility, experiences, and preliminary effectiveness of employing a social robot in this context. The authors reported a statistically significant improvement in participants' sleep hygiene following the robot-assisted educational program.

Alemi et al. [3] utilized a NAO robot to alleviate pain and distress in children with cancer in a hospital setting, observing a considerable reduction in participants' stress, depression, and anger during the robot-assisted program. Prihatini et al. [11] proposed the design of a social robot named Volunteer in pediatric hospitals. The authors claimed that their robot has the potential to interact with children through voice and gestures, assists in assessing the mental health of children with cancer, raises awareness about drug use, and creates enjoyable sounds and music for these children. However, the specific performance results of this robot for children with cancer are yet to be reported and the authors have mainly focused on their robot's design and describing its potentials for individuals with cancer. In addition to these studies, various other robots designed for pediatric hospitals or the education of children with special needs exist. For instance, the Sushi Bot [12], a desk robot with a living face displayed on a screen, was utilized to examine the impact of making eye contact during medical sessions. This study explored social reactions to socially motivating factors, revealing that the presence of a social robot with advanced movements, mimicking human facial expressions, and emotional speech elicited stronger reactions, particularly when users felt the physical presence of the intermediary agent—the robot. Another study investigated how digital games, facilitated by robots, could support cancer patients [13]. In this instance, the Aflaak robot, shaped like a duck, was specifically crafted to provide support and entertainment for children battling cancer as they undergo chemotherapy. The design incorporated stickers, allowing children to engage in playful activities with the robot. As a companion during the challenging chemotherapy process, children could play games with the duck, reducing their pain and making the treatment experience more bearable. The Aflaak robot offered additional features, such as the ability to be dressed up or danced with, contributing to further pain reduction during treatment. Equipped with a microphone and an internal light sensor, the Aflaak robot could perceive its environment and adjust its behavior accordingly, responding to the touch of the children [13]. Additionally, the IROMEC project [14] delved into the role of an automated toy robot in providing medical care and education for children with special needs. This robot came in two configurations—standing and moving. The study involved the design of various scenarios for play and education tailored to the strengths and needs of children with special requirements, such as those with autism. The robot incorporated a monitor to display faces and emotions through images related to different emotions. These images were presented sequentially with gradual changes on the monitor, enhancing the interactive and educational aspects of the robot [14]. Various social robots have been specifically designed to support children in hospital settings, with notable examples being the Arash, Huggable, and Probo robots [15-17]. The conceptual design of the Arash robot for hospitalized children was conducted in 2018 by Meghdari et al. [15]. Acknowledging the challenges children face in harsh hospital conditions, the study aimed to enhance the quality of life for children with cancer. Arash featured four main interaction modules: seeing, hearing, speaking, and gesture. Capable of recognizing users' faces and scanning their surroundings, the Arash robot was designed to move through different hospital areas, interacting with children to educate, motivate, entertain, and alleviate their suffering. The robot comprised three sections—mobile platform, torso, and a suitable physical appearance—with fourteen degrees of freedom, including eight in its arms, two in the neck for realistic head movements, two in the platform, and two in the waist. A 7-inch monitor served as Arash's face [15]. Another social robot tailored for pediatric hospitals is the Huggable robot [16]. Equipped with twelve degrees of freedom, Huggable engaged in play and entertainment for children, featuring touch sensors to facilitate interaction. In the mentioned study, a doctor collaborated with the robot to better support patients. Over 50 hospitalized children were randomly assigned to three groups, each exposed to either a Huggable robotic bear, a virtual Huggable application on a tablet, or an ordinary teddy bear. The researchers reported promising impacts on some children, highlighting the potential usefulness of robotic technology in healthcare applications [16]. In a similar vein, the Probo robot was designed as an animal-shaped assistant for medical staff to entertain sick children in hospitals [17]. Boasting twenty highly accurate motors, Probo could move its ears, eyebrows, eyelids, eyes, trunk, mouth, and neck to enhance interaction with people [17]. These social robots exemplify the potential of robotics in providing emotional support, entertainment, and assistance in healthcare settings, particularly for pediatric patients.

Although different robot-assisted studies have been done for children with cancer so far, there is still a long way to reach an appropriate practical educational/therapeutic solution for these children; so we need to have more studies in this field (while focusing on different items) to see the generalizability, repeatability, and reliability of the previous findings in diverse pediatric hospitals as well as different cultures (especially considering the issue that most of the conducted studies have been done on a few number of children with cancer). Combining positive aspects of previous studies to develop new robot-assisted game-based packages could be going one step further in this line of research.

## C. Main Goals of this Study and the Research Questions

In this paper, we have conceived and crafted Maya, a social robot resembling an elephant, with the dual purpose of serving as a tutor/teacher's assistant and a comforting companion for children. The primary aim is to provide entertainment and education to sick children undergoing treatment in pediatric hospitals. It is noteworthy that the foundational design of the Maya platform was introduced in our prior research [18], encompassing crucial aspects of a social robot and gauging its acceptance among users. Our earlier observations suggested that Maya harbors the potential to serve as an affordable assistive social robot. Building upon this foundation, we have advanced the capabilities of Maya in this paper and performed two pilot studies for children with cancer. Drawing inspiration from the successful design of the Aflaak robot and the noticeable results of its study [13], Maya has been meticulously crafted in the form of a small portable toy animal. Its velvety fabric adds a tactile dimension, providing enjoyment for children as they interact with and embrace Maya. Configured as a smart toy resembling a cuddly baby elephant, the child-friendly Maya aims to bring prolonged comfort to hospitalized children. Crucially, the Maya robot boasts the ability to recognize a spectrum of facial expressions, including sadness, happiness, anger, stress, surprise, disgust, and a neutral face state. This capability facilitates the execution of entertaining and educational scenarios. Furthermore, the inclusion of face recognition allows Maya to recognize individual users across subsequent sessions, tailoring interactions to the unique needs and preferences of each child. Maya represents a technological advancement with the potential to bring joy, comfort, and enhanced educational experiences to children undergoing medical treatment in pediatric hospital settings. The principal contribution of this paper lies in executing meticulously planned experiments targeting a particularly challenging demographic—children with cancer. This group poses substantial accessibility challenges for conducting robot-assisted interventions, requiring adherence to stringent permission protocols that were successfully navigated to obtain approval for the study. The overarching goal of our pilot intervention phase is to yield preliminary exploratory insights into the following two research questions:

(1) Can the attention of children with cancer be effectively diverted from the pain of an injection through robot interaction, ultimately reducing their pain?

(2) What is the perspective of children with cancer and their parents regarding the Maya robot's game-playing capabilities as an entertaining and educational package?

To address these two questions, the experimental phase of our pilot study was bifurcated into two distinct components:

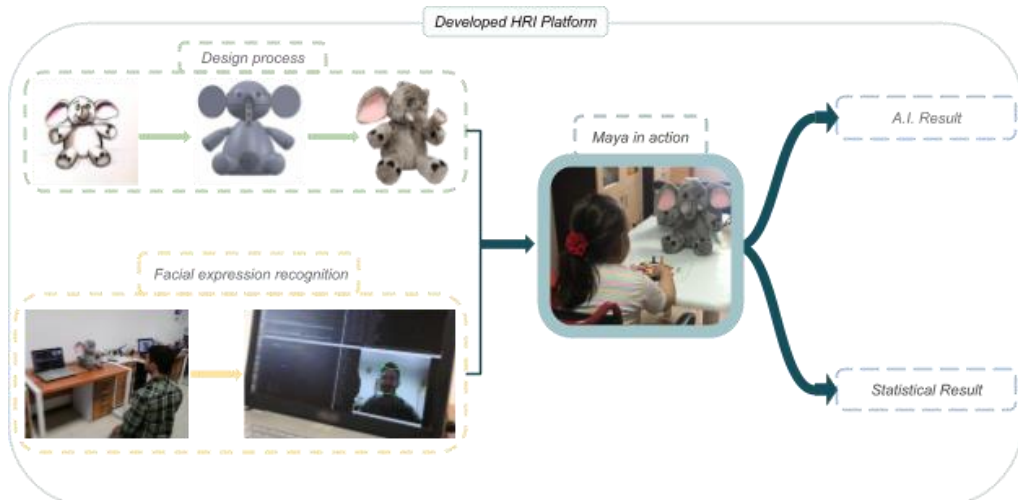

Figure 1: A graphical overview of this study.

(1) Analysis of Pain during Injection Process with Robot Interaction: This segment focuses on quantifying the level of pain experienced by children during the injection process. The robot actively engages with the children by performing amusing scenarios during these interactions. The objective is to assess the initial impact of the robot's intervention on mitigating the perceived pain associated with medical procedures.

(2) Clinical Test with the UTAUT Questionnaire: The second part involves a pilot clinical test that utilizes the Unified Theory of Acceptance and Use of Technology (UTAUT) questionnaire. This questionnaire is employed to gauge the attitudes and perspectives of two specific groups—children with cancer and their parents—while engaging with a robot-based game. Through this exploratory assessment, the study aims to discern the reception and acceptance of the Maya robot in the context of entertainment and education.

As a summary, the design of this study is inspired by the previous similar researches specially the work done with the Aflaak robot [13] and three of our previous works [3, 15, 18]. However, what make this study different in using the proposed robot-assisted game-based package to find an exploratory answers to the mentioned research questions are as follows:

- The shape of the designed/used robot (i.e., an elephant) in this study is based on the selection/viewpoints of some children with cancer among some different suggested animals ([18]).
- We have conducted the pain perception test to initially quantify/compare the perceived pain by the children with/without having a robot in the injection room (that to the best of our knowledge, it has not been conducted in previous studies).
- We tried to provide an educational opportunity similar to the healthy children in our country during a pilot study (i.e., starting to teach the English language to children at that age's range)
- We involved the parents of the children in the technology-based educational/clinical process, too.

For a visual representation of the study's structure and the interplay between these experimental components, refer to Figure 1, which provides a graphical overview of the research methodology. This approach is designed to holistically explore the potential benefits and perceptions associated with robot-assisted interventions for children with cancer.

**II. Design and Development**

Maya, a social toy robot, is meticulously crafted to cater to children dealing with cancer and chronic diseases in both healthcare and educational environments. The design and development of this animal-like social robot align with practical and research goals, encompassing the following specifications.

*A. Design Elements*

Since very past times, dolls and stuffed animals have played a serious role in the childhood or the teenage years. Children are naturally drawn to animal toys [18]. In general, children prefer to interact with their dolls and action figures using body language, verbal communication, and facial expressions, just as they would interact with other children; they also choose names for the human-like or animal-like dolls and play turn-taking game with them. Moreover, due to religious restrictions (for some countries like Muslim population), incorporating real (ritually unclean) animals into daily life is difficult for most Muslims. Therefore, most animal-assisted therapies cannot be performed in Muslim countries such as Islamic Republic of Iran. Social robotics could help bridge this gap in such populations and support the treatment process for patients with disabilities such as cancer [8, 15]. In this context, Maya, resembling a petite toy elephant, serves as a valuable companion for children undergoing extended hospitalization. Beyond its endearing appearance, Maya contributes to therapeutic-educational interactions, providing emotional support and alleviating stress and pain. The design intentionally mirrors that of a regular stuffed toy, sharing dimensions to ensure familiarity and comfort. Drawing inspiration from the developmental influence of animal characters, as highlighted in Walsh's study, exemplified by characters in The Lion King teaching children resilience, Maya inherits this approach to offer solace and serve as positive role models for children in need, functioning as cherished toy friends [19]. It should be noted that any general assumptions such as "real animals are universally attractive" needs further examination/investigation to determine whether it applies across cultures or if it is more context-specific (which is out of the scope of this study).

The design process of Maya prioritized the specific needs and interests of children, taking into careful consideration their unique conditions. Initial discussions with medical staff helped identify the primary requirements, followed by a survey to gather the opinions of the children themselves, ensuring a comprehensive understanding for refining the robot's design. Consequently, Maya was crafted to resemble a toy animal, leveraging its diminutive size to create an ambiance reminiscent of a cherished children's toy, facilitating comfortable and affectionate interactions through hugging.

Maya's design goes beyond a mere physical resemblance, incorporating diverse communication modules tailored to the individual needs of children with cancer. These modules encompass the recognition of seven fundamental emotions expressed through facial expressions, the ability to convey emotions, speech capabilities, responsive body gestures, and the proficiency to recognize and recall different individuals' faces. This multifaceted approach enables Maya to engage with children in a personalized manner, catering to their distinct needs and fostering meaningful interactions.

### B. Dimensions and Degrees of Freedom

The Maya robot benefits from a cost-effective design strategy, incorporating suitable manufacturing processes, notably 3D printing technology. This judicious selection of actuators and sensors based on cost considerations significantly reduces both manufacturing and production costs. The utilization of quick and economical manufacturing methods, particularly 3D printing, not only enhances cost-effectiveness but also facilitates swift and budget-friendly future repair and maintenance procedures. This is particularly crucial for social robots intended for prolonged use in human-robot interaction studies, especially in educational settings.

Designed with practicality in mind, Maya stands at a height of thirty-two centimeters and weighs two kilograms, ensuring ease of transportation. The robot is equipped with a video camera and a microphone situated behind each eye, complemented by a speaker embedded in its chest. Striving for a natural appearance within the constraints of its small size, Maya features five degrees of freedom, distributed across its ears, two hands, trunk, and neck. This streamlined design not only enhances the robot's aesthetic appeal but also optimizes its functionality for diverse applications in human-robot interaction studies.

### C. Motors

The Maya robot incorporates five Dynamixel motors strategically employed for various functionalities. One motor is dedicated to the robot's neck, facilitating head movement and enabling tracking of a child's face after recognition. This dynamic feature enhances the robot's believability, creating a more engaging experience for children. During interactions with the child, the robot utilizes two motors to move its trunk up and down, open and close its ears, and manipulate its hands independently. These expressive movements serve to capture the attention of children, fostering interaction and providing a welcomed distraction from ongoing medical procedures. The elephant-inspired trunk of the robot comprises three distinct moving parts intricately connected to a motor. This design choice ensures a more realistic and captivating movement of the trunk, effectively increasing interest and attention from children participating in hospital experiments (Figure 2-a) [18]. This thoughtful integration of motorized movements enhances the overall interactive experience, contributing to the positive impact of Maya in healthcare settings.

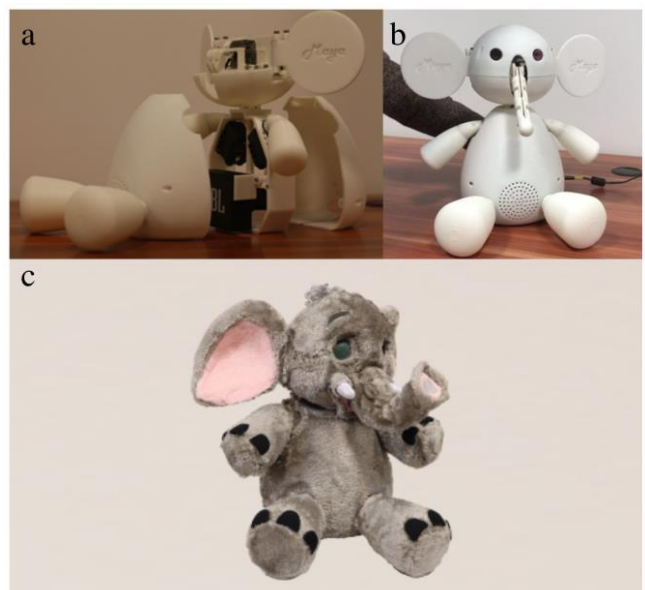

Figure 2. a) The robot's outer shell, b) An external and internal view of the elephant robot, and c) Maya's cloth covering.

### D. Outer cover

Maya's internal components are safeguarded by a robust plastic shell, offering protection against potential impacts from children (Figure 2-b). Enhancing the tactile experience for children, a thick, soft fabric envelope this durable plastic shell, providing a plush and comforting sensation that makes the robot more enjoyable to hug. Prioritizing hygiene in hospital environments, the cloth cover is designed to be both washable and replaceable (Figure 2-c), mitigating the risk of disease transmission and infections. This meticulous attention to material selection ensures not only the durability and safety of the robot but also addresses the unique considerations of healthcare settings where Maya is intended to operate.

### III. Using Artificial Intelligence Algorithms on the Robot

Leveraging artificial intelligence, we have endowed the robot with the capability to recognize faces and interpret facial expressions. The implementation utilizes a ready-made library for face recognition [20], based on the FaceNet network [21]. This integration enables the Maya robot to accurately identify faces, a fundamental feature for effective interactions in social settings. Furthermore, the robot demonstrates the ability to discern the user's facial expressions, a crucial skill for social robots engaging with individuals. The implementation details of the Facial Landmark algorithm, integral to this recognition process, are elaborated upon in the subsequent subsections. This sophisticated integration of AI technologies enhances Maya's social interaction capabilities, making it a more responsive and engaging companion.

#### A. Preprocessing: Face Recognition and its Alignment, using the Openface Algorithm

The utilization of deep neural networks poses a challenge due to the substantial requirement for a large dataset. Collecting appropriate data from human faces becomes crucial, especially considering the varied rotations faces may exhibit in images. To address this challenge, a considerable dataset is essential to effectively train a neural network. One possible solution involves employing the open face algorithm [22], which accurately positions facial components—such as eyebrows, eyes, nose, and mouth—from precise parts of an image. This alignment ensures that these facial elements are consistently placed in specific positions, mitigating the impact of varying orientations and contributing to the robust training of the neural network.

#### B. Features Extraction: Characteristic Points of the Face

Human facial emotions manifest through movements in the eyebrows, eyelids, nose, lips, and variations in skin wrinkling. In this study, the facial landmark method, based on the algorithm presented in [23, 24], was employed to identify components of the human face using 68 key points. This comprehensive algorithm detects all facial components. Recognizing that the lower roundness of the face is less critical for analyzing facial expressions, this element was intentionally excluded from each face, resulting in the featured image as depicted in Figure 3. The method offers advantages such as streamlined data generation, detailed in the subsequent sections, and facilitates the training of a more straightforward neural network.

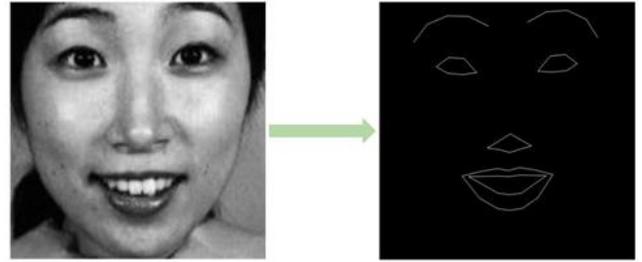

Figure 3. The facial components extracted by the Facial Landmark Algorithm [25].

#### C. Dataset

To compile the datasets essential for facial expression recognition, two widely used datasets, CK+ [25] and MMI [26], were employed. These datasets provide a diverse and comprehensive range of facial expressions, facilitating the robust training and validation of the facial recognition system. The utilization of these well-established datasets contributes to the effectiveness and reliability of the facial expression recognition model implemented in the Maya robot.

1) CK+ Database

The extended version of this dataset encompasses 593 video sequences featuring 123 different individuals. These sequences, ranging from 10 to 60 frames, capture the transition of facial expressions from a neutral state to the final peak expression. The database meticulously labels seven emotional expressions, encompassing the six basic emotional states of sadness, happiness, anger, stress, surprise, and disgust, along with a neutral expression [25]. This comprehensive database serves as a valuable resource for training and evaluating facial expression recognition systems, providing a diverse set of expressions and emotions for robust model development.

2) MMI Dataset

The MMI database comprises facial expressions captured from 20 individuals aged between 19 and 62, with 44% of the participants being female. The sequences within this database systematically present the neutral expression state as the initial frame, followed by each of the six emotional expressions exhibited by the individual's face [26, 27]. This dataset, characterized by its diverse participant demographic, offers a valuable collection of facial expressions for training and assessing the accuracy of facial expression recognition algorithms, enhancing the model's robustness across different age groups and genders.

3) Dataset preparation

For each of the six emotional and neutral facial expressions, 107 images were curated from a combination of the aforementioned datasets, culminating in a total of 749 images. Recognizing that this dataset size might be relatively modest for a deep neural network, a feature extraction process was initiated using the facial landmark method outlined in subsection 3.2. Specifically, the position of point #30 (located centrally) from each image was extracted and utilized to partition the image into two halves. The first half incorporates the eyebrows and eyes, while the second half encompasses the nose and lips (see Figure 4). Subsequently, permutations of all images within each of the seven emotion groups were computed (refer to Figure 5), leading to the generation of new images. This augmentation methodology significantly expands each emotion group to encompass 11,449 images, resulting in a dataset total of 80,143 instances. Following the collection and randomization of the data, the dataset was stratified into three subsets: training data, validation data, and test data, comprising 70%, 10%, and 20% of the dataset, respectively. This systematic approach ensures a robust and diverse dataset for training and evaluating the facial expression recognition model implemented in the Maya robot.

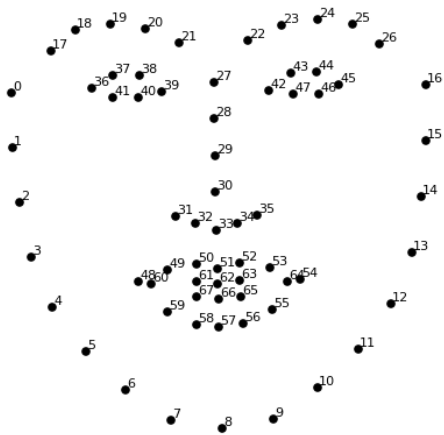

Figure 4. The dots used in the Facial Landmark Algorithm [28].

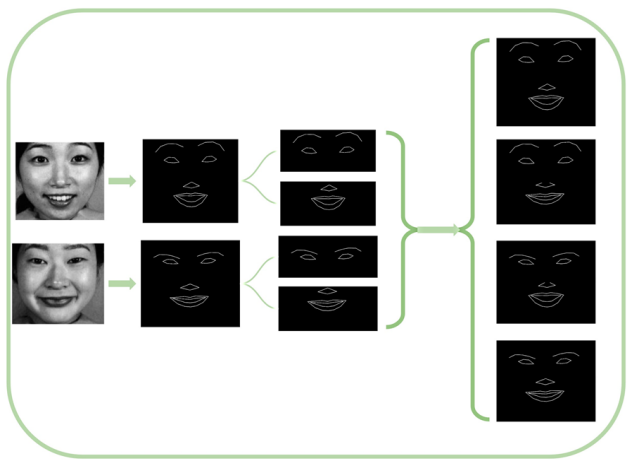

Figure 5. Production of new data for the dataset (Happiness expression) [25].

### D. Deep Network

To facilitate Maya's facial expression recognition, a simple deep neural network was trained employing a methodology similar to that outlined in [29], yielding promising results. In summary, a network was devised incorporating two layers of inception, and the network's architecture is succinctly summarized in Table 1. The ADAM optimizer was employed with a learning rate set at 0.001. Rectified Linear Units (ReLUs) served as the activation function for all layers, while the softmax function was utilized for classifying the seven facial expressions. This well-structured neural network forms the backbone of Maya's facial expression recognition capabilities, demonstrating its effectiveness in accurately identifying and classifying diverse emotional states.

### E. Implementation on the Maya Robot

Maya operates on the Robot Operating System (ROS). To enhance the flexibility of the robot's performance, its software was modularized into smaller components or nodes. Specifically, for facial expression recognition, the program was segmented into four distinct parts: camera initialization, execution of the OpenFace algorithm, generation of the face model, and implementation of the facial recognition algorithm. Following the facial expression recognition process, the robot seamlessly executes predefined scenarios aligned with the recognized emotional expression, in accordance with the study protocols. This modularized approach ensures the efficiency and adaptability of Maya's facial expression recognition capabilities within the broader system architecture.

## IV. Using Artificial Intelligence Algorithms on the Robot

As previously indicated, the Maya social robot was meticulously designed for children within the age range of 4 to 9, with a specific focus on children undergoing cancer treatment. To assess the impact of the Maya robot on these children, and as the main contributions of the paper, two distinct pilot experiments were formulated—one involving music and storytelling, and the other centered around a game. Crucially, both experiments transpired within hospital settings (see Figure 6). These exploratory experiments were strategically devised to furnish initial exploratory insights into the study's overarching research questions. The specific details of the experimental configuration are elucidated in this section. For the sake of clarity, the research questions are restated below:

(1) Can the attention of children with cancer be effectively diverted from the pain of an injection through robot interaction, ultimately reducing their pain?

(2) What is the perspective of children with cancer and their parents regarding the Maya robot's game-playing capabilities as an entertaining and educational package?

Table 1. Network Configuration.

| Layer type | Patch size / Stride | Output size | 1x1 | 3x3 reduce | 3x3 | 5x5 reduce | 5x5 | Pool Proj. | Param. |
|---|---|---|---|---|---|---|---|---|---|
| Convolution -1 | 7x7 / 2 | 48x48x64 | | | | | | | 3.2K |
| Max pool -1 | 3x3 / 2 | 24x24x64 | | | | | | | |
| Convolution -2 | 3x3 / 2 | 12x12x192 | | | | | | | 110.8K |
| Max pool -2 | 3x3 / 2 | 6x6x192 | | | | | | | |
| Inception -3 | | 6x6x32 | 8 | 12 | 16 | 2 | 4 | 4 | 7K |
| Max pool -4 | 3x3 / 2 | 3x3x32 | | | | | | | |
| Inception -5 | | 3x3x50 | 24 | 12 | 12 | 2 | 6 | 8 | 3.1K |
| Avg pool -6 | 3x3 / 1 | 1x1x50 | | | | | | | |
| Convolution -7 | 1x1 / 1 | 1x1x1024 | | | | | | | 52.2K |
| Fully Connected -8 | | 1x1x48 | | | | | | | 49.2K |
| L2 normalization | | 1x1x48 | | | | | | | |
| Total | | | | | | | | | 226K |

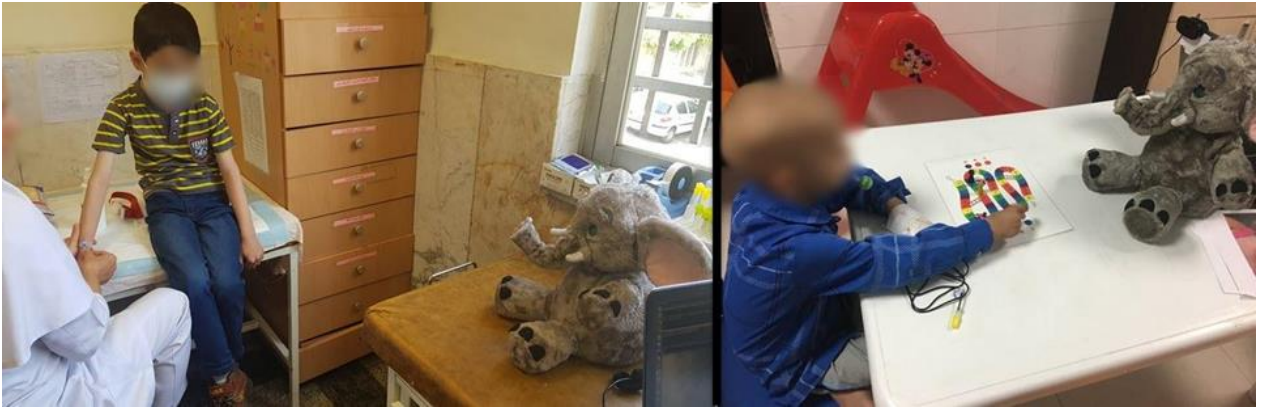

Figure 6. Pictures of the tests being conducted in the hospital.

*A. Participants*

The experiments in this study involved two distinct groups, each serving a specific purpose. The first group was selected to assess pain perception, while the second group was engaged to evaluate child-robot interactions. The initial group comprised 25 children with cancer, aged between 4 and 9, with the distribution of 12 girls and 13 boys. Meanwhile, the second test group encompassed 40 new participants, constituting 20 children with cancer aged 4 to 9 years (including 10 boys and 10 girls) alongside their respective parents. It is noteworthy that, throughout both experiments, the presence of the robot operator, the child's parent(s), and a hospital nurse was maintained in the room. This collaborative involvement ensured a thorough and regulated environment for the experiments. Prior to each experiment, a brief explanation was provided to acquaint both children and their parents with the robot and the procedures involved.

*B. Experimental Protocols*

In this section, we addressed the two main questions of this article. It is noteworthy that in a related study, an emotion and memory model was introduced for a social robot with the capability of creating emotional memories and adapting accordingly. The model was implemented on the NAO robot, using it to teach vocabulary to children through the popular game 'Snakes and Ladders'. Preliminary results revealed that the behavior generated based on this model sustained social engagement and contributed to the improvement of children's vocabulary. The authors also mentioned that the situation where NAO displayed positive emotional feedback had a significantly positive effect on the participant's vocabulary learning performance in comparison to the other conditions of their study [30, 31].

1) Pain Perception (Research Question 1)

To explore the first research question, a preliminary two-mode scenario was devised for the injection process. In mode-A, devoid of the robot, 25 children rated their pain from 0 to 10 after an injection [32, 33]. In mode-B, during another session with the same children, the robot played music and danced during the injection process. Using the Pain Perception Chart (Figure 7), children assigned scores after both sessions. The order of the two modes was randomly assigned to each subject, with mode-A first for 13 participants and mode-B first for the remaining 12 children, creating a counterbalanced condition. All sessions involved the child, parent(s), nurse, and robot operator in the injection room.

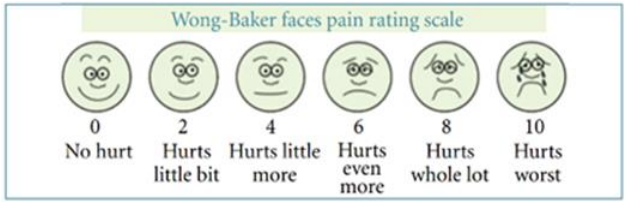

Figure 7. Pain Perception Chart [32, 33].

2) Child-Robot Interaction (Question 2)

In addressing the second research question, an exploratory interactive game between the child and the robot was devised (see Figure 8). The pilot game aimed to foster interaction with both entertainment and educational objectives. To transform the educational atmosphere into a more enjoyable experience, a Snake and Ladder (here Elephant and Ladder!) game was meticulously designed and embellished. The game served the dual purpose of teaching the English language and assessing the robot's proficiency in recognizing facial expressions. This approach to language instruction within a game framework was intended to infuse a sense of joy and eliminate the usual tedium associated with education. As a review, contrary to their healthy peers, children with cancer cannot usually attend foreign language classes to learn a second language (like English) during their hospital treatment periods; therefore, we tried to provide an appropriate educational environment based on their needs on our participants' age range. It should be noted that the designed game was used as a way that the robot can interact with children while showing its capabilities as an assistive tool in educational and entertainment session. Moreover, we considered the point that children receive the most effective/best educational points from those they like (such as the Maya robot); and not necessarily those with the best knowledge/technical skills.

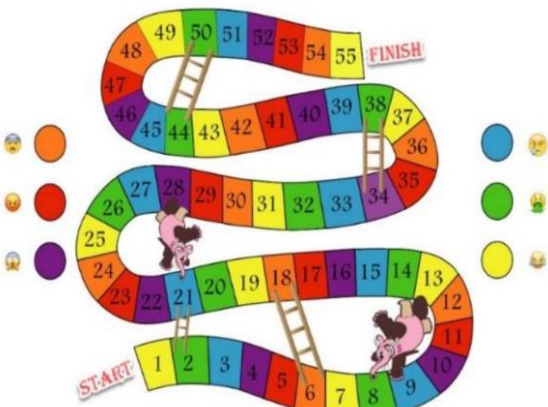

Figure 8. Educational game played by the child and the robot (i.e., the designed Elephant and Ladder game).

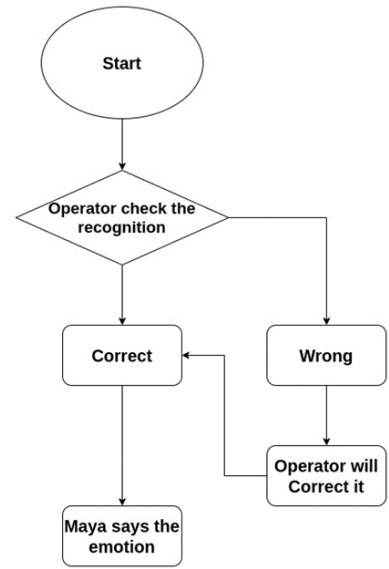

Figure 9. The general flowchart of the performance of facial expression recognition.

Each cell in the game corresponds to a specific emotion, represented by a face emoji for easy recognition during play. The game unfolds as follows: the human mediator explains the rules to the child and instructs them to maintain a neutral facial expression for a few seconds, allowing the Maya robot to recognize both their face and neutral emotional state. The robot then greets the child, asks for their name, and invites them to roll the dice. The child's move is determined by the dice, and based on the color of the cell they land on (each color associated with an emotion), the robot, under the operator's command, states the Persian word for the emotion and prompts the child to provide its English counterpart. If the child is unfamiliar with the English term, the robot introduces the word, instructs the child to repeat it with Maya, and then imitate the corresponding facial expression. This interaction aids the robot in recognizing the child's facial expressions for each emotion. If the child's expression is not accurate, the robot encourages him/her to try again for the game to proceed. During the robot's turn, controlled by the human mediator, the dice is thrown, and Maya's playing piece is moved accordingly. The parent did not play the game and observe the whole process of the child's game. The game aims to teach English words for six emotions while fostering an enjoyable environment for the children. The first player to reach the top cell of the game board is declared the winner, and at the game's conclusion, the robot expresses gratitude and bids farewell to the child. It should be noted that the collected/captured pictures from the participants during the game package has not been published anywhere due to the ethical issue and the consent forms signed by the authors, participants, and the medical staff.

*C. Assessment tools*

1) Pain Perception
Following the pilot implementation of Pain Perception modes-A and B, participants evaluated their pain levels (with a score in the range of 0 to 10) using the provided Pain Perception chart (Figure 7). A paired t-test was employed to assess whether a significant difference exists in the participants' perceived pain between the two conducted modes (refer to the Results section for detailed findings).

2) Child-Robot Interaction
In the exploratory child-robot interaction experiment, we examined potential differences in the attitudes of children and parents towards the robot and the game, utilizing the UTAUT questionnaire [34]. The UTAUT questionnaire is crafted to evaluate the likelihood of success for new technology [34].
A modified version tailored specifically for an assistive social robot has been applied in various studies, including [35]. This questionnaire furnishes insights into various aspects, encompassing Anxiety, Attitude Towards Technology, Facilitating Conditions, Intention to Use, Perceived Adaptability, Perceived Enjoyment, Perceived Ease of Use, Perceived Sociability, Perceived Usefulness, Social Influence, Social Presence, and Trust. The questionnaire utilized in this study is presented in Table 2.

We also added four questions to the ATEG section of the questionnaire specifically to gather the opinions of both children and parents regarding the educational scenario presented by the robot. Additionally, considering that the Maya robot is an animal toy robot, the questions related to humanoid robots were removed from the original version of the questionnaire. Consequently, the total number of questions in our modified questionnaire was 43. Since "the robot and the game" are evaluated as one package, the word "robot" was substituted with "the robot and the game" in the original version of the UTAUT questionnaire, except for the SP and PS categories, which assess the two groups' attitudes toward the robot only (refer to Table 2). Therefore, we used a modified version of the UTAUT in order to preliminary investigate the acceptance of our designed package of robot and the game. Respondents utilized a five-option Likert scale (strongly disagree: 1, disagree: 2, neither agree nor disagree: 3, agree: 4, and strongly agree: 5) to express their attitudes when answering the questions [36, 37]. The scores obtained from the children and their parents are used to find a preliminary answer for the second research question of our study.

**V. Results and Discussion**
This section entails the presentation and discussion of the trained CNN model's performance in facial expression recognition, along with the outcomes of our two experimental interventions aimed at addressing the research questions posed in this study.

*A. Results of the Network for facial expression recognition Points of the Face*
The simple CNN model proposed in this study was trained using the Keras API, demonstrating a classification accuracy of 98% for the seven facial expressions on the dataset. It should be noted that the initial obtained performance of this simple network was fairly promising for conducting our experimental sessions. Just to have more details, Figure 10-a displays the confusion matrix for the test data, consisting of 16,029 photos. It is important to note that the dataset included individuals looking directly at the camera, with controlled lighting conditions and environmental factors. In real-world scenarios, variations in these factors could impact the network's performance. In our experimental settings, which involved 700 annotated photos of 20 children interacting with the Maya robot, the accuracy of the network was observed to be ~79%, as depicted in Figure 10-b. Despite the simplicity of the designed network, it takes approximately 0.3 seconds for Maya's computer to pre-process frames, allowing for video analysis at a rate of three frames per second. Real-time application of trained deep networks remains a common challenge in studies of this nature. While several studies have explored automatic emotion detection, primarily based on deep learning [1, 39], our study focuses on a method capable of real-time performance with reasonable results, considering the hardware restrictions of our robot. It is important to note that the primary contribution of this study lies in conducting designed experiments for children with cancer, with the implementation of the face detection and facial expression analysis system on the Maya robot serving as tools to enhance child-robot interactions and achieve the stated goals (and not necessarily providing new deep networks).

*B. Results of the Pilot Clinical Studies*
In the following two subsections, we present/discuss our lessons learned and the empirical findings of conducting our preliminary exploratory study for children with cancer in the pediatric hospital as the main contributions of the paper. As a review, in the first experiment, we would like to investigate the role of having/not having the Maya robot (while it performed music and dance) on the perceived pain of the children; and in the second experiment, the robot played the emotion game for some new children with cancer and their parents. As mentioned previously, we conducted t-tests for analyzing the results of our exploratory experiments: the pain perception and the educational sessions.

Table 2. The study's modified UTAUT questionnaire.

| Construct | Items |
|---|---|
| Anxiety (ANX) | 1) If I should use the robot and game, I would be afraid to make mistakes with it<br>2) If I should use the robot and game, I would be afraid to break something<br>3) I find the robot and game scary<br>4) I find the robot and game intimidating |
| Attitude towards technology (ATT) | 5) I think it's a good idea to use the robot and game<br>6) The robot and game would make life more interesting<br>7) It's good to make use of the robot and game |
| Facilitating conditions (FC) | 8) I have everything I need to use the robot and game<br>9) I know enough about the robot and game to make good use of it |
| Intention to Use (ITU) | 10) I think I'll use the robot and game during the next few days<br>11) I'm certain to use the robot and game during the next few days<br>12) I plan to use the robot and game during the next few days |
| Perceived Adaptiveness (PAD) | 13) I think the robot and game can be adaptive to what I need<br>14) I think the robot and game will only do what I need at that particular moment<br>15) I think the robot and game will help me when I consider it to be necessary |
| Perceived Enjoyment (PENJ) | 16) I enjoy the robot and game talking to me<br>17) I enjoy doing things with the robot and game<br>18) I find the robot and game enjoyable<br>19) I find the robot and game fascinating<br>20) I find the robot and game boring |
| Perceived Ease of Use (PEOU) | 21) I think I will know quickly how to use the robot and game<br>22) I find the robot and game easy to use<br>23) I think I can use the robot and game without any help<br>24) I think I can use the robot and game when there is someone around to help me<br>25) I think I can use the robot and game when I have a good manual |
| Perceived Sociability (PS) | 26) I consider the robot a pleasant conversational partner<br>27) I find the robot pleasant to interact with<br>28) I feel the robot understands me<br>29) I think the robot is nice |
| Perceived Usefulness (PU) | 30) I think the robot and game are useful to me<br>31) It would be convenient for me to have the robot and game<br>32) I think the robot and game can help me with many things |
| Social Influence (SI) | 33) I think the staff would like me to use the robot and game<br>34) I think it would give a good impression if I should use the robot and game |
| Social Presence (SP) | 35) It sometimes felt as if the robot was really looking at me<br>36) I can imagine the robot to be a living creature<br>37) Sometimes the robot seems to have real feelings |
| Trust | 38) I would trust the robot and game if it gave me advice<br>39) I would follow the advice the robot and game give me |
| Attitude towards Educational Game (ATEG) | 40) I pay attention to the educational content that the robot and game are teaching me<br>41) Education is interesting when using the robot and game<br>42) I think using the robot and the game for education is more effective than normal education<br>43) I find education boring when using the robot and game |

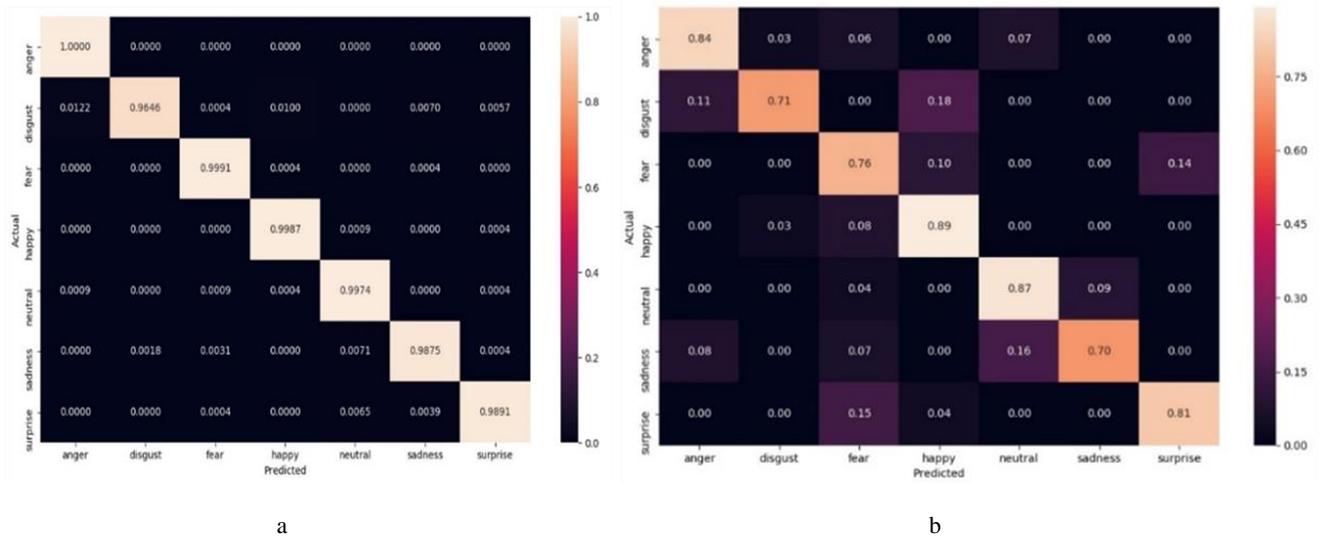

Figure 10. a) The Confusion Matrix of the test set, b) The Confusion Matrix of the children's data in this study.

Table 3. The mean score, standard deviation, and p-value of children and parents from the modified UTAUT questionnaire. P-values are less than (or around) 0.05 are shown in bold.

| No. | Item | Score's mean[a] (SD) | | P-value |
|---|---|---|---|---|
| | | **Children** | **parents** | |
| 1 | ANX | 1.63 (0.64) | 1.68 (0.72) | 0.81 |
| 2 | ATT | 4.73 (0.48) | 4.37 (0.68) | **0.06** |
| 3 | FC | 3.88 (0.84) | 4.15 (0.61) | 0.25 |
| 4 | ITU | 4.50 (0.51) | 4.28 (0.75) | 0.28 |
| 5 | PAD | 4.17 (0.80) | 4.28 (0.61) | 0.62 |
| 6 | PENJ | 4.66 (0.49) | 4.70 (0.50) | 0.80 |
| 7 | PEOU | 4.06 (1.03) | 4.34 (0.65) | 0.31 |
| 8 | PS | 4.66 (0.65) | 4.43 (0.67) | 0.27 |
| 9 | PU | 4.45 (0.62) | 4.40 (0.64) | 0.80 |
| 10 | SI | 4.33 (0.69) | 4.28 (0.63) | 0.81 |
| 11 | SP | 4.41 (0.67) | 4.13 (0.64) | 0.18 |
| 12 | Trust | 4.73 (0.45) | 4.33 (0.61) | **0.02** |
| 13 | ATEG | 4.83 (0.41) | 4.64 (0.51) | 0.20 |

SD: Standard deviation
[a] Score's mean is out of 5

Table 4. The mean score, standard deviation, and p-value of children and parents scores from selected questions. P-values are less than 0.05 are shown in bold.

| Questions | Score's mean[a] (SD) | | P-value |
|---|---|---|---|
| | Children | Parents | |
| Q6 | 4.65 (0.57) | 4.20 (0.75) | **0.04** |
| Q7 | 4.80 (0.40) | 4.40 (0.58) | **0.01** |
| Q26 | 4.80 (0.51) | 4.35 (0.73) | **0.03** |
| Q36 | 4.50 (0.59) | 3.80 (0.68) | **0.001** |
| Q42 | 4.90 (0.30) | 4.75 (0.54) | 0.28 |
| Q43 | 1.10 (0.30) | 1.35 (0.48) | **0.05** |

SD: Standard deviation
[a] Score's mean is out of 5

*1) Pain Perception*

To analyze the pain perception test data, a paired t-test was employed to compare the pain scores of children during injections in modes-A and B, with and without the presence of the robot. The results revealed a significant difference ($p < 0.001$) between the mean pain scores of the children in the absence or presence of the robot. The mean scores were 8.56 and 4.60 (out of 10) for modes-A and B, respectively. This suggests that the presence and activities of the Maya robot were effective in reducing the pain experienced by this group of 25 children (Figure 11).

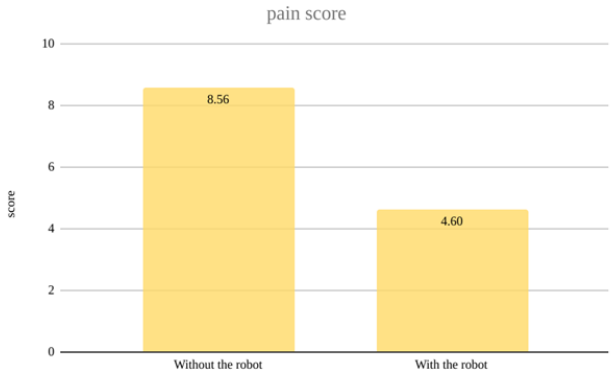

Figure 11: The Pain Report chart for the participated children.

In a study similar to ours, researchers in [40] utilized a NAO robot to alleviate anxiety and distract participants during vaccine injections. Their analysis of the Wong-Baker Scale scores via a paired t-test revealed a significant difference between the two groups.

*2) Child-Robot Interaction*

Table 3 illustrates the results of the t-test comparing the two groups of children and parents across 13 categories of the UTAUT questionnaire in our second pilot experiments. The findings indicated a significant difference between the two groups only in the Trust category (with $p < 0.05$), indicating that parents exhibited less trust in the robot and the game than their children.

Trust in robots, as emphasized in previous studies [41, 42], plays a crucial role in shaping the human-robot relationship, and this extends to trust in the game. Louie [43] also highlighted the importance of socially assistive robots reducing client anxiety during the initial encounter to enhance the user's ease of using the robot. Additionally, the p-value in the Attitude category could be considered a marginally significant difference between the groups.

From the perspectives of both children and parents, the ANX item received a low score, indicating that participants felt calm during human-robot interactions while playing the game. Fortunately, children felt quite relaxed during the playing game with Maya. Consequently, we can reach the following conclusion: Regarding having the parents in the experimental room, we posit that observing a child's joy could potentially alleviate maternal stress. Spending time in the child's playroom, or engaging with the child, even briefly, may enhance the parent-child relationship. Consequently, the mother's sense of usefulness may improve as she assists her child in completing tasks accurately. Additionally, the child's satisfaction derived from participating in the joyful robot-assisted program might contribute to increased compliance with parental directives across various domains, consequently enhancing the mother's positive emotions. This observation is supported by the notion that active parental involvement in the treatment process leads to the children applying acquired skills at home and in other contexts [8, 15, 45]. The high scores on FC, ITU, and PEOU items suggest that both groups felt capable of interacting with the robot and playing the game appropriately. It is worth noting that there was no significant difference between the two groups in any of the items. Further analysis of specific questions on the questionnaire revealed a significant difference between the two groups for questions 6 and 7, indicating that the children's group had a more positive attitude towards using the robot and the game compared to their parents. This aligns with the positive attitude reported towards the social robot role and patients in previous research, leading to higher motivation and better performances in treatment protocols during robot-assisted interventions. Additionally, a significant difference was observed between the two groups (children and parents) in question 26 of the PS item and question 36 of the SP item, both evaluating Maya's social interaction. The results suggest that the robot's social relationship was perceived more positively among children than among parents. This observation is consistent with previous research indicating that real or virtual social robots are generally more appealing to children than their parents during human-robot interactions. Another noteworthy finding is the responses to question 42, determining the attitude towards the effectiveness of education using the robot and game compared to regular education classes. Although there was no significant difference in the answers between the two groups, their high scores suggest a belief that the education provided by the robot and the game is effective. This aligns with previous research highlighting the potential of social robots in achieving outcomes similar to human tutoring and improving cognitive and emotional outcomes in education. The significant difference observed in question 43 indicates that education by the robot and game is found to be more interesting for the children in this group than for their parents. This resonates with the notion that social robots are seen as fun and motivating for children, and parents appreciate the qualities of social robots, such as patience and non-judgmental assistance, in helping their children.

*C. Summary of the Results and Discussion*

Maya's primary goals revolve around education and promoting healthy behaviors. The intelligent toy is equipped with multimedia devices and exhibits various behaviors, including self-presentation. Social robots, like Maya, face critical evaluations related to their acceptance and usability. To assess these aspects, a pilot scenario involving a poem, story, and music was designed. The findings suggest that Maya holds potential in therapeutic applications, demonstrating the ability to be utilized for entertainment and educational purposes for children. The scenario's positive results indicate that Maya can contribute to enhancing the well-being and engagement of its users.

Based on our obtained findings in this preliminary exploratory study, we can conclude that the Maya robot has the ability to reduce the perceived pain of the children during the injection process as well as it does have the ability to educate the participant in pediatric hospitals.

One of the interesting lessons learnt from this study was from a brief interview with one of the nurses responsible for injections after our experiments: She mentioned that "the Maya robot's cute appearance is interesting, and such robots could assist nurses in their job. Given the nature of our work, children often harbor negative feelings towards us/nurses because of doing the injections (despite from how kind we/nurses are!). These robots could be beneficial in various situations, such as education and reducing anxiety in the hospital environment".

## VI. Limitations and Future Work

Due to the unique challenges associated with children with cancer, including limited accessibility and time constraints, enrolling additional volunteers or extending the interaction duration was not feasible in this study. Future research endeavors should focus on enhancing the usability and control of the robot, particularly for healthcare staff and non-professional operators, such as parents. This improvement can be achieved through the design of user-friendly Graphical User Interfaces (GUIs). Additionally, the collection of new datasets from children's faces (which is a confidential not-publishable dataset) in hospital settings, utilizing modern AI algorithms like meta-learning and continual learning, can contribute to refining the robot's capabilities. To broaden the applicability of the robot, efforts should be directed towards making it more versatile and suitable for various educational environments, ranging from schools and kindergartens to home settings. Incorporating voice recognition and action recognition features will enable the robot to better understand and respond to children's voices and movements, enhancing the overall interactive experience. While the feasibility of using the social robot in a pediatric hospital has been demonstrated, the next steps involve implementing diverse educational programs to enhance the knowledge and well-being of children with cancer through interactions with the Maya robot. It is essential to continue advancing the capabilities of intelligent systems in healthcare, ensuring they are explainable and trustworthy, as highlighted by the researchers in [48, 49].

It is important to highlight that the observed outcomes in the second experiment of this study are influenced by simultaneous factors: a) the presence of robots in the sessions, and b) the nature of the designed educational game, and c) the present of the parents. It should be considered that we could not distinguish the individual effects of these factors on the children's performances. To conduct a comprehensive foundational analysis of the relative effectiveness of the developed game versus the designed robot, future research might involve replacing the robot with cartoon character or human in the same intervention scenarios.

## VII. Conclusion

The Maya robot, equipped with intelligent features such as face recognition and facial expression recognition, serves a crucial purpose in creating positive interactions with children, particularly those dealing with cancer, during medical treatments. In this research, during a preliminary exploratory study, two clinical tests were conducted within a pediatric hospital, focusing on the dual goals of entertaining children to reduce pain during injections and assisting in their education. The study delved into the development of interaction and communication between children and the robot during the educational games. Moreover, the article provided insights by comparing the perspectives of parents and children regarding the utilization of the Maya robot in the treatment process. The preliminary findings suggest a generally positive impact of the robot on children with cancer, indicating the feasibility of employing the social robot in a pediatric hospital setting. By conducting this study, we have observed some promising exploratory results and some interesting learned lessons.

## ACKNOWLEDGMENT
This study was funded by the "Dr. Ali Akbar Siassi Memorial Research Grant Award" and the Sharif University of Technology (Grant No. G4030507). Cooperation of Ali Asghar and Mofid Hospitals staff in preparing the ground for experimentations is appreciated. We also thank Mrs. Shari Holderread for the English editing of the final manuscript.

## Conflict of interest
Author Alireza Taheri has received a research grant from the Sharif University of Technology (Grant No. G4030507). The authors Minoo Alemi, Elham Ranjkar, Raman Rafatnejad, and Ali F. Meghdari assert that they have no conflict of interest.

## Availability of data and material (data transparency)
All data from this project (videos of the sessions, results of the questionnaires, etc.) are available in the archive of the Social & Cognitive Robotics Laboratory, Sharif University of Technology.

## Code availability
All of the codes are available in the archive of the Social & Cognitive Robotics Laboratory, Sharif University of Technology. In case the readers need the codes, they may contact the corresponding author.

## Authors' contributions

All authors contributed to the study's conception and design. Material preparation, data collection, and analysis were performed by Elham Ranjkar and Raman Rafatnejad. The first draft of the manuscript was written by Alireza Taheri, Elham Ranjkar and Raman Rafatnejad. Alireza Taheri, Minoo Alemi, and Ali F. Meghdari have supervised the study. All authors read and approved the final manuscript.

## Ethical Approval

Ethical approval for the protocol of this study was provided by the Iran University of Medical Sciences (#IR.IUMS.REC.1395.95301469).

## Consent to participate

Informed consent was obtained from all individual participants included in the study.

## Consent for publication

The authors affirm that human research participants provided informed consent for the publication of the subjects' images. All of the participants have consented to the submission of the results of this study to the journal.

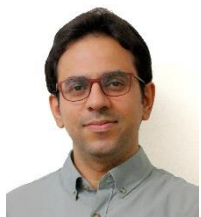
**Alireza Taheri** is an Associate Professor of Mechanical Engineering with an emphasis on Social and Cognitive Robotics at Sharif University of Technology, Tehran, Iran. He is the Head of the Social and Cognitive Robotics Lab. and the Measurement Systems Lab. at Sharif University of Technology. The line of his research focuses on designing/using Social and Cognitive Robotics, Virtual Reality Systems, and Human-Robot Interaction (HRI) platforms for education and rehabilitation of children with special needs (e.g. children with autism, children with hearing problems, children with cerebral palsy). His researches include robots' design and fabrication, serious games' design, artificial intelligence and control, conducting educational/clinical interventions for children, developing cognitive architectures for social robots, mathematical modeling of participants' behaviors during HRI, and empowering robots to analyze users' behaviors automatically and then react adaptively.

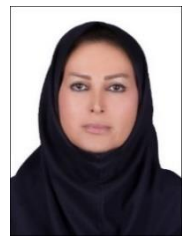
**Minoo Alemi** received her Ph.D. in Applied Linguistics from Allameh Tabataba'i University in 2011. She is currently a Professor and Division Head of Applied Linguistics at the Islamic Azad University, West-Tehran Branch. She is the co-founder of Social Robotics in Iran, a title she achieved as a Post-Doctoral research associate at the Social Robotics Laboratory of the Sharif University of Technology. Her areas of interest include discourse analysis, interlanguage pragmatics, materials development, and RALL. Dr. Alemi has been the recipient of various teaching and research awards from Sharif University of Technology, Allameh Tabataba'i University, Islamic Azad University, and Int. Conf. on Social Robotics (ICSR-2014).

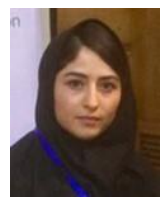
**Elham Ranjkar** received her M.S. degree in Mechanical Engineering from Islamic Azad University, Tehran, Iran. Her research interests include Mechatronics, Artificial Intelligence, and Machine Learning.

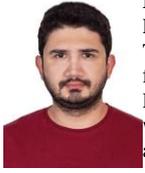

**Raman Rafatnejad** received his B.S. degree in Robotics Engineering from the Islamic Azad University, North Tehran Branch. He was involved in the development of the first generation of social robots at the Social and Cognitive Robotics Lab at Sharif University of Technology, Tehran, where he focused on robot design, computer vision, artificial intelligence and human-robot interaction.

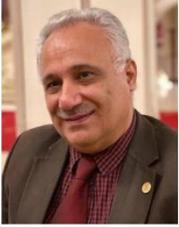

**Ali F. Meghdari** is a Professor Emeritus of Mechanical Engineering and Robotics at Sharif University of Technology (SUT) in Tehran. Professor Meghdari has performed extensive research in various areas of robotics; social and cognitive robotics, mechatronics, and modeling of biomechanical systems. He has been the recipient of various scholarships and awards, including: the 2012 Allameh Tabatabaei distinguished professorship award by the National Elites Foundation of Iran (BMN), the 2001 Mechanical Engineering Distinguished Professorship Award from the Ministry of Science, Research & Technology (MSRT) in Iran, and the 1997 ISESCO Award in Technology from Morocco. He is the founder of the Centre of Excellence in Design, Robotics, and Automation (CEDRA), an affiliate member of the Iranian Academy of Sciences (IAS), a Fellow of the American Society of Mechanical Engineers (ASME), and the Founder and Chancellor of Islamic Azad University- Fereshtegaan International Branch (for students with special needs; primarily the Deaf).